\documentclass[10pt,journal,cspaper,compsoc]{IEEEtran}

\usepackage{times}
\usepackage{epsfig}
\usepackage{epstopdf}
\usepackage{graphicx}
\usepackage{amsmath}
\usepackage{amssymb}
\usepackage{multirow}
\usepackage{float}
\usepackage{algorithm} %format of the algorithm
\usepackage{algorithmic} %format of the algorithm
\usepackage[colorlinks,urlcolor=black,linkcolor=black,anchorcolor=black,citecolor=black]{hyperref}

\begin{document}

\title{Semantic tracking: Single-target tracking with inter-supervised convolutional networks}
\author{Jingjing~Xiao,~\IEEEmembership{Member,~IEEE,}
        Qiang~Lan,
        Linbo~Qiao,
        Ale\v{s}~Leonardis,~\IEEEmembership{Member,~IEEE}% <-this % stops a space
\IEEEcompsocitemizethanks{\IEEEcompsocthanksitem J. Xiao and A. Leonardis are with the University of Birmingham, United Kingdom, E-mail: shine636363@sina.com,  a.leonardis@cs.bham.ac.uk
\IEEEcompsocthanksitem L. Qiao and Q. Lan are with the College of Computer, National University of Defense Technology, China, E-mail: \{lanqiang, qiao.linbo\}@nudt.edu.cn
}}

% \markboth{IEEE Transactions on Pattern Analysis and Machine Intelligence}%
% {\MakeLowercase{\textit{J. Xiao et al.}}:Semantic tracking: Single-target tracking with inter-supervised convolutional networks}

\IEEEtitleabstractindextext{%
\begin{abstract}
\renewcommand{\raggedright}{\leftskip=0pt \rightskip=0pt plus 0cm} This article presents a semantic tracker which simultaneously tracks a single target and recognises its category. In general, it is hard to design a tracking model suitable for all object categories, e.g., a rigid tracker for a car is not suitable for a deformable gymnast. Category-based trackers usually achieve superior tracking performance for the objects of that specific category, but have difficulties being generalised. Therefore, we propose a novel unified robust tracking framework which explicitly encodes both generic features and category-based features. The tracker consists of a shared convolutional network (NetS), which feeds into two parallel networks, NetC for classification and NetT for tracking. NetS is pre-trained on ImageNet to serve as a generic feature extractor across the different object categories for NetC and NetT. NetC utilises those features within fully connected layers to classify the object category. NetT has multiple branches, corresponding to multiple categories, to distinguish the tracked object from the background. Since each branch in NetT is trained by the videos of a specific category or groups of similar categories, NetT encodes category-based features for tracking. During online tracking, NetC and NetT jointly determine the target regions with the right category and foreground labels for target estimation. To improve the robustness and precision, NetC and NetT inter-supervise each other and trigger network adaptation when their outputs are ambiguous for the same image regions (i.e., when the category label contradicts the foreground/background classification). We have compared the performance of our tracker to other state-of-the-art trackers on a large-scale tracking benchmark~\cite{wu2015object} (100 sequences)---the obtained results demonstrate the effectiveness of our proposed tracker as it outperformed other 38 state-of-the-art tracking algorithms.
\raggedright
\end{abstract}

\begin{IEEEkeywords}
Single-target tracking, convolutional networks, semantic tracking 
\end{IEEEkeywords}}

\maketitle

% \IEEEpeerreviewmaketitle

\section{Introduction}\label{SecIntro}

% 1. topic; 2.exiting problems; 3. promising method
Visual object tracking has actively been researched for several decades. Depending on the prior information about the target category, the tracking algorithms are usually classified as {\em category-free methods}, like KCF~\cite{henriques2015high}, Struck~\cite{hare2011struck}, LGT~\cite{Cehovin2013a}, and {\em category-based} methods, like human tracking~\cite{vondrak2013dynamical}, vehicle tracking~\cite{Cao2014168}, hand tracking~\cite{oikonomidis2012tracking}. The category-free tracking methods are acknowledged for their simple initialisation (a single bounding box) and easy generalisation across different object categories. They have extensively been studied and compared~\cite{wu2015object, VOT2014}. However, as those methods have no prior information about the {\em target} inside the bounding box, the tracking performance heavily depends on the heuristic assumptions of image regions, i.e., appearance consistency~\cite{XIAO2015CVPR} and motion consistency~\cite{chi2014novel}, which fail when those assumptions are not met. In contrast, the category-based methods benefit from the prior information about the target and can better adjust the target model and predict its dynamics or appearance variations during tracking. Those category-based methods can achieve superior performance on a specific category but usually have difficulties being generalised to other object categories. 
As many sophisticated machine learning algorithms have recently been adopted for tracking~\cite{Ma2015ICCV, wangvisual, wang15}, an interesting question is whether we can build a~\textit{semantic tracker}, based on those methods, to bridge the gap between the category-free tracking methods and category-based tracking methods (see Tab.~\ref{TabIntro}). Early attempts to track and recognise the objects simultaneously were done by~\cite{lee2005visual,fan2013we,yun2016kernel}. However, the aforementioned works were developed using conventional hand-crafted features, which have difficulties of being scaled up. Inspired by the recent success of convolutional networks~\cite{krizhevsky2012imagenet}, we propose, in this article, a~\textit{semantic tracker} with a unified convolutional framework which encodes generic features across different object categories while also captures category-based features for model adaptation during tracking. With the help of the {\em category-classification network}, the semantic tracker can avoid heuristic assumptions about the tracked objects.

 %The relationships among category-free, category-based and the proposed semantic tracking methods are shown in Tab.~\ref{TabIntro}. where to build a unified framework to train the target model across different object categories and use this information to simultaneously recognise and track the unknown target on-line

\begin{table*}[ht]
\centering
\caption{Relationships among category-free, category-based methods and the proposed semantic tracking. Category-based methods and the proposed semantic tracking encompass off-line category-specific training processes whereas the category-free methods do not. During online tracking, only the category-based methods know the target category from the initialisation stage while the proposed semantic tracking algorithm simultaneously recognises and tracks the target on-the-fly.}
\label{TabIntro}
\begin{tabular}{|c|c|c|c|}
\hline
\multirow{2}{*}{Methods} & \multirow{2}{*}{Off-line category-specific training} & \multicolumn{2}{l|}{Online tracking: target category} \\ \cline{3-4} 
                         &                           & Initialization            & Output             \\ \hline
Category-free tracker & No                        & Unknown                   & Unknown            \\ \hline
Category-based tracker& Yes                       & Known                     & Known              \\ \hline
Proposed semantic tracker & Yes                       & Unknown                   & Known              \\ \hline
\end{tabular}
\end{table*}

% Note that semantic tracking is not simply combining the recognition results with the category-based tracking methods. Besides the similar spatial semantic information in recognition or classification, we explicitly designed a temporal semantic model to interpret the target dynamics or appearance variation during tracking which avoids the heuristic assumptions about the target. 

\begin{figure}[t]
\centering
\includegraphics[width = 0.5 \textwidth]{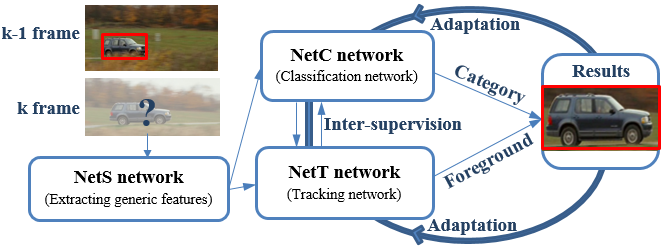}
\caption{The architecture of the proposed semantic tracker, which contains a shared convolutional network (NetS), a classification network (NetC) and a tracking network (NetT).}
\label{FigIntro}
\end{figure}

The proposed semantic tracker comprises three stages: off-line training, online tracking, and network adaptation. It consists of a shared convolutional network (NetS), a classification network (NetC) and a tracking network (NetT), see Fig.~\ref{FigIntro}. In the off-line training stage, NetS is pre-trained from ImageNet to extract generic features across different object categories. Those features are then fed into NetC for classification and NetT for tracking. Note that NetT has multiple branches to distinguish the tracked object from the background. Since each branch is trained by the videos of a specific object category, this enables each branch in NetT to learn the category-specific features related to both foreground and background, e.g., when tracking a pedestrian, it is more likely to learn the features of a car in the background than features of a fish. 
During online tracking, NetC first recognises the object category and activates the corresponding branch in NetT. Then, NetT is automatically fine-tuned for that particular tracking video by exploiting the foreground and the background sample regions in the first frame. When a new image frame arrives, the algorithm samples a set of image regions and each sample is fed through both NetC and NetT. The regions with the right category and the foreground label are used for target estimation (i.e., the location and the size of the target bounding box). Note that the target appearance often changes during the tracking, therefore it is extremely crucial for a tracker to adapt the model accordingly. To improve the robustness and precision, NetC and NetT inter-supervise each other and trigger network adaptation when their outputs are ambiguous (i.e., not consistent) for several image regions, % (i.e., when the category labels from NetC contradict the foreground/background classification from NetT; 
e.g., when an image region is classified as a non-target category from NetC but as foreground from NetT or as a target category from NetC and background from NetT. The samples with consistent labellings are used to update the networks which also results in a reduced number of ambiguous sample regions. We have evaluated the contribution of each key component to the overall performance on OTB tracking benchmark~\cite{wu2015object} (100 sequences), and also compared the whole algorithm to the other state-of-the-art single-target tracking algorithms. The experimental results demonstrate the effectiveness of our algorithm as it outperformed other 38 state-of-the-art tracking algorithms not only overall, but also on the sub-datasets annotated with specific attributes.

Different from conventional category-free and category-based trackers, the main contributions of our semantic tracker can be summarised as: 

\begin{enumerate}
\item Our tracker simultaneously tracks a single target and recognises its category using convolutional networks, which alleviates the problems with heuristic assumptions about the targets; 
\item A novel unified framework with NetS network, which extracts generic features across different object categories, combined with NetC and NetT networks which encode category-based features;
\item NetC and NetT jointly determine image samples for estimation of the target, and inter-supervise each other by triggering network adaptation to improve robustness and precision. 
\end{enumerate}

The rest of the paper is organised as follows. We first review related work in Sec.~\ref{SecRW}. The details of the proposed method are provided in Sec.~\ref{SecPT}. Sec.~\ref{SecER} presents and discusses the experimental results on a tracking benchmark~\cite{wu2015object}. Sec.~\ref{SecCon} provides concluding remarks.

% The experimental results show that the proposed algorithm can: 1) recover from short-term recognition failure during tracking; 2) transfer the semantic information to different object categories; 3) achieve the state-of-the-art tracking performance.

\section{Related work}\label{SecRW}
% category-free tracking: single cue->multi cues->semantic information
% We can review this history in a perspective of feature abstraction, which is used as the theoretical base of Deep Learning.
% category-based tracking
Conventional tracking algorithms can be classified as category-based trackers and category-free trackers. Category-based tracking is targeted at some particular applications, e.g., Vondrak et al.~\cite{vondrak2013dynamical} tracked a human body by considering physical plausibility, Oikonomidis et al.~\cite{oikonomidis2012tracking} tracked a hand with 26-DOF hand model, where Newtonian physics was applied to approximate the rigid-body motion dynamics. %~\cite{oikonomidis2011full} % by explicitly exploiting physical constraints,Pfister et al.~\cite{pfister2012automatic} presented a hand tracker that detected joint positions. 
The mentioned works demonstrate that prior information about the target can significantly help the tracking algorithms to achieve more accurate and robust results. However, the existing category-based (articulate/rigid/dynamic) models and corresponding (physical/common-sense) constraints often suit that particular category and have difficulties being generalised. In contrast, category-free tracking is acknowledged for its simple initialisation (one bounding box) and easy generalisation across different object categories, as has extensively been demonstrated in~\cite{wu2015object, VOT2014}. Early category-free trackers~\cite{nummiaro2003adaptive, mei2009robust, comaniciu2000real, adam2006robust} built the methods on a single feature, which is prone to failure when the applied feature endures large variations. To alleviate the problems of using a single feature, later works~\cite{wu2015robust,wang2015inverse,XIAO2015CVPR,li2016deeptrack} adaptively fused multiple features using sophisticated machine learning algorithms to build a target model to achieve robust tracking. However, in general, it is hard to design a model suitable for all different object categories, e.g., a rigid tracker for a car is not suitable for a deformable gymnast. Therefore, semantic information about the target category becomes essential to enable a tracker to optimize the model during tracking. 

Recent works~\cite{wangvisual, Lebeda15, Ma2015ICCV} began to exploit intrinsic information about the tracked objects, with an attempt to overcome the semantic gap and assist in developing robust tracking algorithms. Lee et al.~\cite{lee2005visual}, Fan et al.~\cite{fan2013we} and Yun and Jing~\cite{yun2016kernel} tried to track and recognise the objects simultaneously, however, these works were based on hand-crafted features, which hampered them to be scaled-up. 

Inspired by the recent success of convolutional networks, Wang et al.~\cite{wangvisual} conducted an in-depth study on the properties of convolutional neural network features (CNN)~\cite{krizhevsky2012imagenet} which showed that the top layers encode more semantic features and serve as category detectors, while lower layers carry more fine-grained details and can better discriminate the target from the background. Therefore,~\cite{wangvisual} jointly used those layers with a switch mechanism during the tracking. A similar work was done by Ma et al.~\cite{Ma2015ICCV}, where they exploited CNN features~\cite{simonyan2014very} trained on ImageNet~\cite{deng2009imagenet} to improve tracking accuracy and robustness. Different from~\cite{wangvisual}, where the tracking algorithm was switching between the layers with semantic information and fine-grained information,~\cite{Ma2015ICCV} fused features from hierarchical layers to conduct a coarse-to-fine tracking strategy. However, both trackers,~\cite{Ma2015ICCV,wangvisual}, were off-line pre-trained on ImageNet images~\cite{deng2009imagenet} and then directly used for on-line tracking, without any online fine-tuning of the network structure for a specific tracking task. The realisation that purely using target images for training is not optimal since a target in one video can be part of the background in another, let to the use of videos to train the trackers. Wang et al.~\cite{wang2015video} pre-trained a two-layer CNN based tracker from video sequences, and proposed a domain adaptation method which effectively adapted the pre-learned features according to the specific target during online tracking. Wang et al~\cite{wang2016stct} also proposed a sequence-trained network with generic feature extraction layers from VGG network~\cite{simonyan2014very} and two-layer adaptation network. A similar work was done by Nam et al.~\cite{MDcnn}, who also proposed a video-trained CNN network with a shared network and multi-branches to distinguish the object from the background. However, all the mentioned video-trained trackers~\cite{wang2015video,wang2016stct,MDcnn} did not explicitly exploit the semantic information of the target, i.e., object category. Without knowing the category of the object, it is highly probable that the tracker will learn false positives, and will have difficulties recovering from the failures. In addition, the afore mentioned trackers triggered the network adaptation in a heuristic way with pre-defined time intervals, causing inadequate adaptation which potentially resulted in either model drifting or outdated models. In contrast, our proposed semantic tracker significantly deviates from the aforementioned related works in several aspects including the network structure, initialisation procedure, target estimation and online adaptation, summarised as: 1) we clearly define the shared network NetS for extraction of generic features, followed by the networks NetT and NetC for category-based features extraction. This also brings more intuitive understanding about what we have learnt in each network part; 2) NetT is explicitly trained with multiple branches encoding category-based features, where the corresponding branch is activated by classification network NetC; 3) the samples for the target estimation are jointly decided by the outputs from both NetC and NetT; 4) the network adaptation of NetC and NetT is conducted in an inter-supervised manner when their outputs for the same image region are in contradiction, i.e., a sample is classified by NetT as foreground but not correctly recognised by
NetC or vice-versa;  this  step ensures a proper network updating pace, avoiding heuristics; 5) the proposed work simultaneously tracks the target and recognises its category. 

\section{The proposed tracker}\label{SecPT}
In this section, we first introduce the structure of the proposed tracker model (Sec.~\ref{SecTM}). Then, we explain the off-line training process, which constructs the tracker using ImageNet~\cite{deng2009imagenet} and tracking videos~\cite{VOT2014} (Sec.~\ref{SecOT}). The network intialisation, target estimation and network online adaptation are explained in Sec.~\ref{SecOn}.

\subsection{Tracker model}\label{SecTM}
\begin{figure*}[htb]
\centering
\includegraphics[width = 0.9 \textwidth]{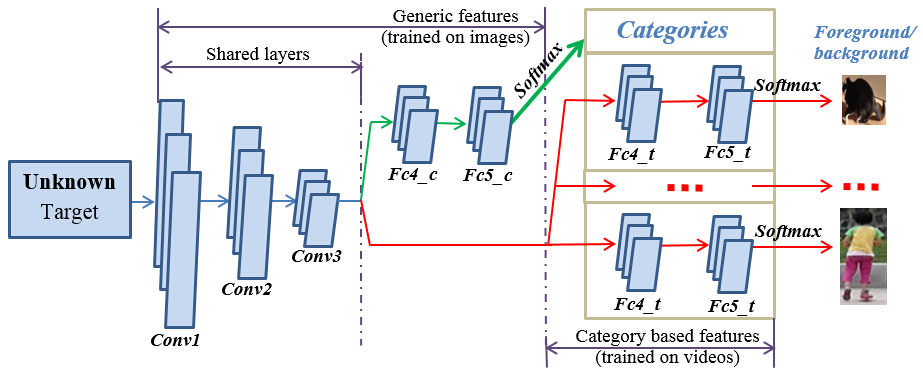}
\caption{The architecture of the proposed semantic tracker, which contains a shared convolutional network (NetS), providing inputs to two networks (NetC and NetT) with fully connected layers. The green arrows indicate NetC for categorising the tracked object. The red arrows indicate NetT for tracking, which comprises multiple branches, and each branch is particularly trained for specific object categories.}
\label{Fig2}
\end{figure*}

% explain the model and why we define it in such a way
Recent research has shown the relationship between the human vision system and deep hierarchies in computer vision~\cite{kruger2013deep}. CNNs, being partly inspired by these ideas, are acknowledged for their outstanding representation power and have extensively been studied in~\cite{krizhevsky2012imagenet,simonyan2014very}. Therefore, we also build our semantic tracker based on CNN components, but propose a new architecture illustrated in Fig.~\ref{Fig2}.

Recent research~\cite{Ma2015ICCV} has shown that shallow layers in CNN contain more generic information while deep layers are more related to semantic information. Thus, our tracker consists of shared convolutional layers to extract generic features in the shallow network (NetS), followed by NetC network for classification and NetT network for extracting category-based features for tracking. Note that NetS extracts generic features across different object categories, where those features have some common properties, e.g., robustness to scale and orientation changes, and illumination variations~\cite{MDcnn}, which can be useful for other higher level tasks. Therefore, those extracted generic features are fed into NetC and NetT for more semantic related tasks. NetC is a multi-class classification network to recognise the object category. NetT, which is a binary classification network, aims at distinguishing foreground region (target) from the background. Considering that the images of tracked objects of the same category often contain characteristic features both in terms of the foreground as well as the background, but which are different from other categories, e.g., when tracking a pedestrian it is more likely to have cars in the background than fish, NetT comprises multiple category-based branches, and each branch is particularly trained from the videos that contain the same object category. During on-line tracking, NetC and NetT inter-supervise each other by triggering network adaptation to improve robustness and precision, shown in Fig.~\ref{FigIntro}. The details of the network structure are shown in Tab.~\ref{TabCNN}. 

\begin{table*}[htb]
\centering
\caption{The structure of the proposed semantic tracker. In the convolutional layers, the first number indicates the receptive field size as ``num x size x size'', followed by the convolution stride ``str.'', spatial padding ``pad'', local response normalisation ``lrn'', and the max-pooling down-sampling factor.}
\label{TabCNN}
\begin{tabular}{|c|c|c|c|c|c|c|}
\hline
Structure    & \multicolumn{3}{c|}{~\textbf{NetS shared network}}                                                                                                                                                                                                                                                       & \multicolumn{3}{c|}{~\textbf{NetC network}}                                 \\ \hline
\multirow{5}{*}{\begin{tabular}[c]{@{}c@{}}Input\\ images\\ (107*107)\end{tabular}} & conv1                                                                                             & conv2                                                                                          & conv3                                                                                & fc4\_c                  & \multicolumn{2}{c|}{fc5\_c}     \\ \cline{2-7} 
                                                                                     & \multirow{4}{*}{\begin{tabular}[c]{@{}c@{}}64*11*11,\\ str.4, pad 0,\\ lrn, *2 pool\end{tabular}} & \multirow{4}{*}{\begin{tabular}[c]{@{}c@{}}256*5*5\\ str.1, pad 2\\ lrn, *2 pool\end{tabular}} & \multirow{4}{*}{\begin{tabular}[c]{@{}c@{}}256*3*3\\ str.1,\\ pad 1\end{tabular}} & 256, relu, dropout & \multicolumn{2}{c|}{8, soft-max} \\ \cline{5-7}  & & & & \multicolumn{3}{c|}{~\textbf{NetT network}} \\ \cline{5-7} 
 & & & & fc4\_t & fc5\_t \\ \cline{5-7} 
 & & & & 256, relu, dropout & 2, soft-max    \\ \hline
\end{tabular}
\end{table*}

\subsection{Off-line training}\label{SecOT}
\textbf{NetS for generic features extraction.} 
With extensive CNN-based studies for object classification, several representative models have been proposed and made publicly available, e.g., AlexNet~\cite{krizhevsky2012imagenet}, GoogleNet~\cite{szegedy2015going}, VGGNet~\cite{simonyan2014very} etc. Rather than training the model from scratch, we transfer knowledge from a pre-trained model into NetS to extract generic features. A pre-trained model VGG-f~\cite{Chatfield14} is explicitly chosen, because 1) it is trained from a tremendous dataset ImageNet~\cite{deng2009imagenet}; %2) has relative small network which is preferable for the tracking task~\cite{MDcnn}; 
2) it achieves comparable performance with the fastest speed~\cite{MatConvNet}. Our NetS has the same structure as the first three convolutional layers in VGG-f~\cite{Chatfield14} except that the input image size is adapted (107*107). Since our training dataset is substantially smaller than ImageNet, the shared convolutional layers (NetS) are kept fixed to avoid the over-fitting problem.

\textbf{NetC for classification.}
% training images
NetC aims at recognising the object's category with two fully connected layers. When training NetC with our dataset, NetS first extracts generic features and those features are then fed into NetC network for fine-tuning. Note that the object in the video often undergoes significant deformations and suffers from a poor field of view and partial occlusions. In addition, the generated image samples during tracking might only cover the target partially or the target is not centralised inside the bounding boxes. Therefore, to improve the performance of our classification network NetC, we also prepared training samples with noisy bounding boxes, denoted as:
%Training is performed using object regions $\{X^{n_c}_{k}\}_{n_c=1...N_c}$ extracted from the training images.  To 

\begin{equation}
X^{n}_{c,k}= X_{k} + \Delta X_{c,k}^{n}
\end{equation} 

\noindent where $X_{k}$ is the target ground truth at $k$-th frame, and $\Delta X^{n}_{c,k}$ is the perturbation of the $n$-th sampled region $X^{n}_{c,k}$. Specifically, we generated 50 object samples with significant overlap ratio (0.8) with the ground truth bounding boxes from each frame. To balance the distribution of different target status, those samples are shuffled during training. Note that NetC is trained as a multi-classification network to classify the object regions into different categories by Stochastic Gradient Descent (SGD) method with the learning rate 0.0001 and 128 batch size. The objective function for training NetC network is denoted as:

\begin{equation}
<\hat{W}_{c}, \hat{B}_{c}> = \arg\min \frac{1}{N_c}\sum_{n=1}^{N_c}|| f_c(X_{c, k}^{n})- l_{c, k}^{n}||_2
\end{equation} 

\noindent where $\hat{W}_{c}$ and $\hat{B}_{c}$ are the weights and biases of the NetC network, and $f(X_{c,k}^{n})$ is the predicted label while $l_{c,k}^{n}$ is the ground truth label of the $n$-th image region $X_{c,k}^{n}$ at frame $k$.

\textbf{NetT for tracking.} 
% general information
NetT is a binary classification network with multiple branches corresponding to different object categories, aiming at distinguishing the foreground (object) image regions from the background image regions. Note that the object in one video might become background in another video, but videos belonging to the same category share some intrinsic category-based features in both foreground and background.
Therefore, the category-based branch in NetT can extract the target features with discriminative semantic information. In NetT, each branch has two fully connected layers to further process the generic features from NetS. In each frame of the training videos, we use the same training samples in NetC as positive (target) samples for NetT to preserve training consistency. Beside the positive samples that are the same as used in NetC training, we also generate 200 samples with overlap ratio below 0.2 as negative (background) samples for the training. NetT is trained to classify the positive object regions from negative object regions also using SGD method with the learning rate 0.0001 and 128 batch size, where the learnt weights are denoted as $<\hat{W}_{t},\hat{B}_{t}>$. The whole process of the training procedure is explained below:

\vspace{12pt}
\begin{table}[htb]
\centering
\begin{tabular}{l}
\hline
\textbf{Algorithm 1}: off-line training~~~~~~~~~~~~~~~~~~~~~~~~~~~~~~~~~~~~~~~\\ \hline
1:~\textbf{Input}: the categorised training sequences from VOT benchmark~\cite{VOT2014}. \\ 
2: Prepare the training dataset $\{X^{n}_{c,k}\}_{n=1...N_c}$ for NetC (50 samples\\~~ each frame) and $\{X^{n}_{t, k}\}_{n=1...N_t}$ for NetT (50 positive samples and\\~~ 200 negative samples per frame).\\ 
3: Shuffle the whole NetC training dataset, and the NetT training datasets.\\
4: Train the NetC with the NetC training dataset by SGD, where the low\\~~~ level features are extracted from NetS.\\
5: Train the multi-branch NetT network with the NetT training datasets\\~~~ by SGD, where the low level features are also extracted from NetS.\\
6:~\textbf{Output}: the weights and bias $<\hat{W}_{c}$, $\hat{B}_{c}>$ for the trained 
 NetC network\\~~~ and $<\hat{W}_{t},\hat{B}_{t}>$ for the NetT network.\\\hline
\end{tabular}
\end{table}

\subsection{Online tracking}\label{SecOn}

During the online tracking stage, the algorithm first takes several image regions around the target's position in the previous frame, and feeds them into our network to estimate the target's bounding box. NetS extracts the low-level generic features for NetC and NetT. Then NetC and NetT jointly determine the image regions for target estimation, and inter-supervise each other while updating. 

\textbf{Initialisation.} Given a bounding box in the first frame, we apply the pre-trained NetS and NetC to assign the content of the bounding box to the corresponding NetT branch. To improve the recognition accuracy, we sample the image regions closely around the ground truth (0.8 overlap). If the majority of bounding boxes have the same category label, that category will be regarded as the true object category and activate the corresponding branch in NetT. Note that the same type of the target (e.g., a car) can appear different in different videos, thus we need to fine-tune the activated branch in NetT for a particular tracking video. Therefore, the algorithm samples the image regions around the target for training based on the overlap with the ground truth. For positive (foreground) samples, we initially select 500 image regions with the overlap over 0.8 in the first frame. % Then, the samples which are recognised as the target category by NetC will be used as positive samples. 
For negative (background) samples, we initially select 5000 image regions with the overlap below 0.2. Those samples, classified as other categories, will be treated as negative samples. The generated foreground and background samples are used to fine-tune NetT at the first frame through 30 iterations with the learning rate 0.001.

To improve the tracking accuracy, we need to train the model to estimate the size of the target and adjust the bounding box scale. This is achieved by learning the correspondence between the extracted features and the target size. Recent detection works~\cite{girshick2014rich,redmon2015you} have explored the regression capabilities of the rich hierarchical features, which separate the tasks of associating category probabilities and bounding boxes estimation. Inspired by those regression-based object detectors, we apply the same regression technique~\cite{girshick2014rich} (derived from~\cite{felzenszwalb2010object}) to estimate the scale of the bounding boxes during tracking, aiming at improving the tracking accuracy. To obtain the linear functions $g_x(.), g_y(.), g_w(.), g_h(.)$ that map the features extracted from NetS to the bounding box centre (identified with subscripts $x$ and $y$) and scale (subscript $w$ is width and $h$ is height), we train the bounding box regressors in the first frame as:

\begin{equation}
\label{EqReg}
\left\{
\begin{array}{rl}
g_x(NetS(X_{1}^n)) =&(X_{1,x} - X_{1,x}^n)/X_{1,w}^n\\
g_y(NetS(X_{1}^n)) =&(X_{1,y} - X_{1,y}^n)/X_{1,h}^n\\
g_w(NetS(X_{1}^n)) =&log(X_{1,w}/X_{1,w}^n)\\
g_h(NetS(X_{1}^n)) =&log(X_{1,h}/X_{1,h}^n)
\end{array} \right.
\end{equation}

\noindent where $X_{1,x}$, $X_{1,y}$, $X_{1,w}$, and $X_{1,h}$ are the center ($x$ and $y$ axis coordinates), width and height of the ground truth bounding box $X_{1}$ at the first frame, while $X_{1,x}^n$, $X_{1,y}^n$, $X_{1,w}^n$, and $X_{1,h}^n$ are the corresponding values of the generated bounding box $X_{1}^n$. $NetS(X_{1}^n)$ denotes the features extracted from NetS. To learn the transformation from the generated bounding box to the ground truth bounding box, 10.000 samples are generated and the linear functions are learnt by least squares estimates. During online tracking, those learnt bounding box regressors will be used to improve the bounding box scale estimation every frame.

% JINGJING STOPS HERE: 31/08/2016
\textbf{Semantic tracking.} From the second frame onwards, the algorithm generates $N_f$ ($N_f=256$) candidate image regions subjected to a Gaussian distribution around the previous target position, denoted as:
\begin{equation}
X^n_k=\hat{X}_{k-1} + \Delta X_k^n
\label{EqSamples}
\end{equation} 

\noindent where $\hat{X}_{k-1}$ is the estimated target position at $k-1$ frame, and $\Delta X_k^n$ is the perturbation of the sampled region $X^n_k$. $\Delta X_k^n~\sim~\mathcal{N}(0, R)$ is a zero-mean Gaussian noise with a constant variance-covariance matrix $R$. % {\color{red} Do I understand this correctly that we sample around the previous location in all directions equally. This means that we do not take into account the temporal information (i.e., how the target moves), but that is ok.}

Then, the tracker extracts generic features from each sample by NetS, and feeds those features into NetC for the classification (to determine the category) and NetT for the tracking (to determine foreground/background), denoted as:

\begin{equation}
\left\{
\begin{array}{rl}
f_c(X^n_k) :& NetS(X^n_k) \to NetC \\
f_t(X^n_k) :& NetS(X^n_k) \to NetT 
\end{array} \right.
\end{equation}

\noindent where $f_c(X^n_k)$ is the output of the image sample $X^n_k$ from NetC network, and $f_t(X^n_k)$ is the output of NetT network. Note that no matter how the target appearance changes, the category of the object should remain the same. Therefore, after NetC classifies the samples and assigns them category labels, only the samples labelled as the original category will be treated as potential target samples. The value of $f_c(X^n_k)$ is $1$ when the recognised content of the bounding box is consistent with the active branch in NetT. If not, the value becomes $0$. The value of $f_t(X^n_k)$ ranges between $0$ and $1$, which denotes the likelihood of the sample being a foreground sample. Since NetC and NetT simultaneously classify each sample, there are four different combinations of labels which guide the further process, shown in Tab.~\ref{TabType}.

\begin{table*}[htb]
\centering
\caption{Possible outcomes based on the results of NetC classification network (original/other object category) and NetT tracking network (foreground/background) of each sample.}
\label{TabType}
\begin{tabular}{|c|c|c|c|}
\hline
Sample   & NetC     & NetT       & Outcome                                                                             \\ \hline
Type I   & Original & Foreground & For target estimation; For online updating (a positive sample) \\ \hline
Type II  & Original & Background & \multirow{2}{*}{An ambiguous sample}           \\ \cline{1-3}
Type III & Other  & Foreground &                                                                                         \\ \hline
Type IV  & Other  & Background & For online updating (a negative sample)                                \\ \hline
\end{tabular}
\end{table*}
Samples classified as the original category from NetC and foreground from NetT are regarded as type I samples. Since type I samples obtain consistent (positive) labellings from NetC and NetT, they are regarded as highly trustable target samples and are used to estimate the target, defined as:
\begin{equation}
\hat{X}_k^n = \arg\max f(X^n_k),~~f(X^n_k)=f_c(X^n_k) f_t(X^n_k)
\label{EqTypeI}
\end{equation}

Note that, to improve the robustness of the tracker, instead of using the sample with the highest score in Eq.~\ref{EqTypeI}, we choose $N_{top}$ samples with highest scores for bounding boxes regression. The bounding box regressors learnt in the initialization stage (Eq.~\ref{EqReg}) are applied to estimate the object scale from selected $n$-th image region $\hat{X}_k^n$.
\begin{equation}
\label{EqReg2}
\left\{
\begin{array}{rl}
\tilde{X}_{k,x}^{n} =&g_t(NetS(\hat{X}_{k,x}^{n}))\hat{X}_{k,w}^{n}+\hat{X}_{k,x}^{n}\\
\tilde{X}_{k,y}^{n} =&g_t(NetS(\hat{X}_{k,y}^{n}))\hat{X}_{k,h}^{n}+\hat{X}_{k,y}^{n}\\
\tilde{X}_{k,w}^{n} =&exp(g_t(NetS(\hat{X}_{k,w}^{n})))*\hat{X}_{k,w}^{n}\\
\tilde{X}_{k,h}^{n} =&exp(g_t(NetS(\hat{X}_{k,h}^{n})))*\hat{X}_{k,h}^{n}
\end{array} \right.
\end{equation}

\noindent where subscripts $x, y, w, h$ have the same meaning as in Eq.~\ref{EqReg} for the selected bounding box $\hat{X}_{k}^{n}$ at frame $k$. The final estimation of the target $\hat{X}_k$ utilises the expectation operator over the rescaled samples $\tilde{X}_{k}^{n}$ computed by Eq.~\ref{EqReg2}, denoted as:
\begin{equation}
\hat{X}_k = \frac{1}{N_{top}}\sum_{n=1}^{N_{top}} f(\tilde{X}_{k}^{n})\tilde{X}_{k}^{n}
\label{EqFinal}
\end{equation}

\noindent where $f(\hat{X}_{k}^{n})$ is the score computed from Eq.~\ref{EqTypeI}. $N_{top}$ is the number of selected Type I samples with highest scores. 

\textbf{Inter-supervised network adaptation.} 
To handle appearance variations of the target during tracking, it is important to be able to update the NetC and NetT networks accordingly. There are two essential questions about the network adaptation: 1) when to update and 2) how to update. Ideally, NetC and NetT should obtain consistent conclusions about the same image region, that means that a foreground region should also have the right category label. If not, such ambiguous situations indicate that NetC and NetT need to be re-trained with the newest samples, at which point the network adaptation is triggered.

Note that the type IV samples (the same as the type I samples in Tab.~\ref{TabType}) also obtain consistent labellings (in the case of the type IV they are negative) from both networks. Those samples with consistent labellings are used for later network adaptation when ambiguities occur as a result of NetC and NetT outputting contradictory results (type II and type III samples). As shown in Tab.~\ref{TabType}, the algorithm detects ambiguous samples (AS) when inconsistent labellings arise from the outputs of NetC and NetT, i.e., type II and type III samples. An increasing number of AS indicates that the current networks have difficulties consistently classifying the incoming samples and should be updated. Since NetC is not thoroughly pre-trained with fine-grained information, it may misclassify the object under some (new) conditions. Also, the initially trained foreground/background boundary of NetT may not be reliable any more. Therefore, both NetC and NetT need to be updated with the most recent consistent samples. To update the networks, NetC and NetT use the consistent samples during the process, i.e., type I and type IV samples. While it is straightforward to use type I and type IV samples to update NetT, type IV samples do not have a validated category label to train a specific category in NetC. Therefore, type I samples are used to train the original category in NetC while type IV samples are used to train the category X (unknown category, explained in Sec.~\ref{SecData}) to update NetC, denoted as:

\begin{equation}
\left\{
\begin{array}{rl}
<\hat{W}_{c}, \hat{B}_{c}> =\arg\min \frac{1}{N_{tr}}\sum_{n=1}^{N_{tr}}|| f_c(X_{tr,k}^{n})- l_{c,k}^{n}||_2\\
<\hat{W}_{t}, \hat{B}_{t}> =\arg\min \frac{1}{N_{tr}}\sum_{n=1}^{N_{tr}}|| f_t(X_{tr,k}^{n})- l_{t,k}^{n}||_2
\end{array} \right.
\end{equation}

\noindent where $<\hat{W}_{c}, \hat{B}_{c}>$ and $<\hat{W}_{t}, \hat{B}_{t}>$ are the weights and biases of NetC and NetT,  $\{X_{tr,k}^{n}\}_{n=1...N_{tr}}$ are the type I and type IV samples used for training, $l_{c,k}^{n}$ and $l_{t,k}^{n}$ are the corresponding ground truth labels. After one round of adaptation, the updated NetC and NetT will jointly be used to classify the ambiguous samples again. The newly classified type I or IV samples originating from previous AS will be added into the training pool for the next training iteration. It is expected that the newly trained networks NetC and NetT will produce increasingly consistent labellings for the image regions, which indeed happens, as the number of ambiguous samples is reduced by updated networks. Therefore, we use this as the stopping criterion for the adaptation, i.e., when the number of AS stops decreasing or is sufficiently small (0.2 in practice). The process of online tracking is explained below: 

\begin{table}[H]
\centering
\begin{tabular}{l}
\hline
\textbf{Algorithm 2}: online tracking \\ \hline
1:~\textbf{Input}: the ground truth of the target in the first frame.\\ 
2: Initialise the tracker by recognising the target's category with NetC,\\~~~~activating corresponding branch in NetT and fine-tuning the NetT\\~~~~network with image regions.\\
3: Train the bounding box regressors, Eq.~\ref{EqReg}.\\
4: \textit{For} frame = 2: $N_f$\\
5:~~~~~~Generate candidate images samples with respect with Eq.~\ref{EqSamples}\\
6:~~~~~~Categorise each sample with NetC network and classify the samples\\~~~~~~~~~into the foreground and background with NetT network.\\
7:~~~~~~Choose image samples in terms of Eq.~\ref{EqTypeI} for estimation.\\
8:~~~~~~Estimate the target position and scale, Eq.~\ref{EqReg2}, Eq.~\ref{EqFinal}.\\
9:~~~~~~Calculate the number of AS samples $N_{AS}$.\\
10:~~~~~\textit{While} $N_{AS}>$ threshold\\
11:~~~~~~~~~~~~Fine-tune the NetC and NetT with type I and type IV samples.\\
12:~~~~~~~~~~~~Categorise each sample with NetC network and NetT network.\\
13:~~~~~~~~~~~~Calculate the number of AS samples $N_{AS}$.\\
14:~~~~~~\textit{End}\\
15: \textit{End} \\
16:~\textbf{Output}: the estimated object position and scale.\\
\hline
\end{tabular}
\end{table}

\section{Experimental results}\label{SecER}
In this section, we first explain the implementation details of the tracker. Then, we evaluate the tracker from four aspects: the effectiveness of the tracker sub-components, the qualitative performance compared to other CNN-based trackers, the quantitative performance compared to all other state-of-the-art trackers and the failure cases of the proposed tracker. Finally, we present some ideas for future work~\footnote{The code will be released upon acceptance of the paper.}.

\subsection{Implementation details}
In this section, we provide the details about the datasets, evaluation metrics, as well as training and running speed. 

\subsubsection{Datasets}\label{SecData}
\textit{Training dataset} - To train the tracker we use the sequences from~VOT~\cite{VOT2014}, explicitly excluding the sequences that also appear in OTB~\cite{wu2015object}, which is used as the test dataset. The training dataset was, for the purpose of constructing NetT branches, classified into 8 categories according to the tracked objects, namely, pedestrians, faces, cars, animals, balls, motorbikes, dolls and a category~X (which comprises of the targets that do not fall into any of the 7 categories). 

\textit{Test dataset} - The algorithm is tested on a large scale tracking benchmark OTB~\cite{wu2015object} which has 100 sequences, and each sequence has several tracking attributes to facilitate evaluation. The features of the training dataset and the test dataset are listed in Tab.~\ref{TabData}. 

\begin{table*}[htb]
\setlength{\tabcolsep}{5pt}
\small
\centering
\caption{The features of the training dataset and the test dataset. The training dataset is obtained from VOT~\cite{VOT2014}, explicitly excluding test sequences. The test dataset~\cite{wu2015object} consists of 100 sequences.}
\label{TabData}
\begin{tabular}{|c|c|c|c|c|c|c|c|c|c|}
\hline
\multicolumn{2}{|c|}{Categories}                                                        & Pedestrians & Faces  & Cars   & Animals & Balls & Motorbikes& Dolls & Category~X \\ \hline
\multirow{2}{*}{\begin{tabular}[c]{@{}c@{}}Training\\ set\end{tabular}} & No.of Seq     & 17         & 3     & 6     & 13     & 4    & 3         & 1    & 11     \\ \cline{2-10} 
                                                                        & No. of frames & 5975       & 441   & 3216  & 5412   & 949  & 695       & 326  & 3110   \\ \hline
\multirow{2}{*}{\begin{tabular}[c]{@{}c@{}}Test\\ set\end{tabular}}     & No.of Seq     & 36         & 23    & 12    & 5      & None & 2         & 7    & 15     \\ \cline{2-10} 
                                                                        & No. of frames & 16258      & 11306 & 11223 & 1705   & None & 392       & 8893 & 9263   \\ \hline
\end{tabular}
\end{table*}

\subsubsection{Evaluation metrics} 
We report the results of one pass evaluation (OPE) based on the evaluation protocol proposed in OTB~\cite{wu2015object}. Note that there are two criteria used in the OTB, namely overlap and centre-error. In our experiment, we only use the overlap (success plot) rather than the centre-error (precision plot) in tracking evaluation since the centre distance is: 1) susceptible to subjective bounding box annotations; 2) unreliable in cases when a tracker completely loses a target~\cite{vcehovin2015visual}. Therefore, we use the area under curve (AUC) of the success plot to rank the trackers. 

The overlap ${\varphi _k}$ at frame $k$ is defined by using the tracker-output bounding box $\hat{X}_k$ and ground-truth bounding box $X_k^G$ in Eq.~\ref{EqAcc}: 

\begin{equation}
{\varphi _k} = \frac{|\hat{X}_k  \cap X_k^G|}{|\hat{X}_k  \cup  X_k^G|}
\label{EqAcc}
\end{equation}

\noindent where $ \cap $ and $ \cup $ represent the intersection and union of two regions and $| \bullet |$ is the region size measured by pixels number. 

In the success plot, the x-axis depicts a set of thresholds for the overlap to indicate the tracking success. The success ratio is the number of correctly tracked frames divided by the total number of frames for a more comparable evaluation, Eq.~\ref{EqThresh}.

\begin{equation}
P_{\tau}(\hat{X}_k, X_k^G)=\frac{||\{k|{\varphi _k} >\tau\}_{k=1}^{N_f}||}{N_f}
\label{EqThresh}
\end{equation}

\noindent where $\tau$ denotes the threshold of the overlap, and $N_f$ is the total number of frames. A failure is detected when the overlap (computed in Eq.~\ref{EqAcc}) is below the defined threshold $\tau$. 

\subsubsection{Speed}
The proposed algorithm was implemented in Matlab2014a (linked to some C components) using an Intel i7-4710MQ CPU and Nvidia Quadro K1100M GPU, giving the average training speed of 289.5 bbps (bounding boxes per second) and the test speed of 189.2 bbps. 

\subsection{Evaluation of the sub-components of the tracker}
\begin{figure}[ht]
\centering
\includegraphics[width=3in]{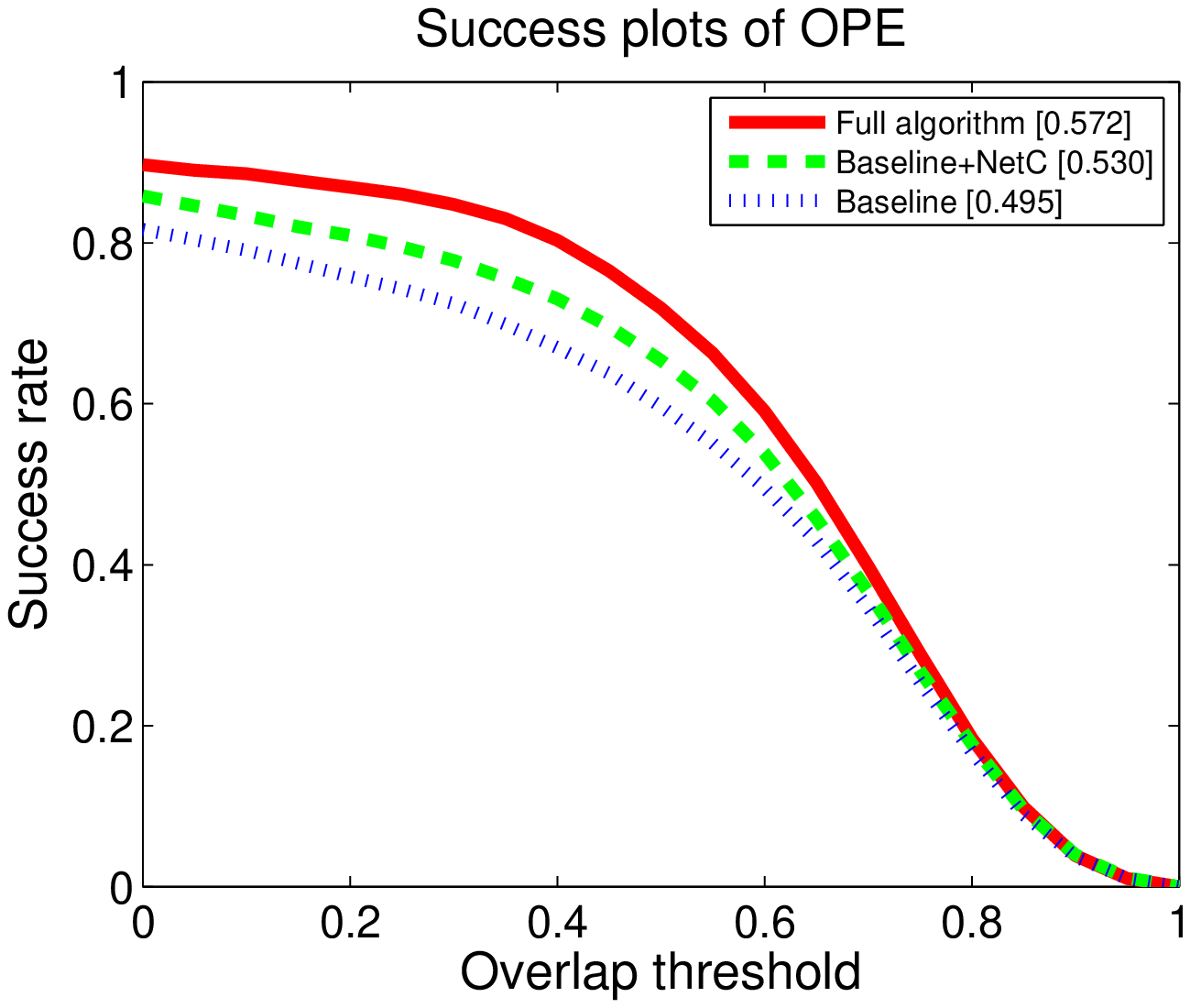}
\includegraphics[width=2.5in, height = 0.3 \textwidth]{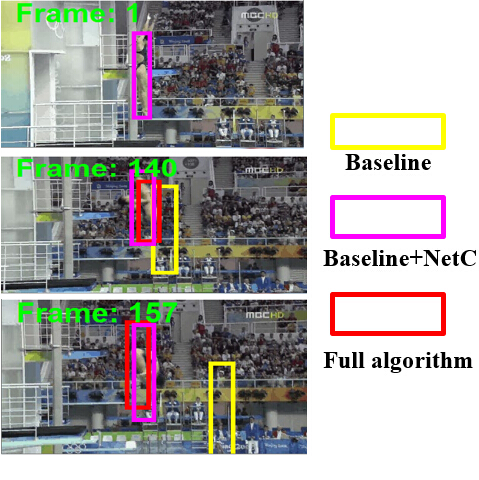}
\caption{Top: Evaluation of the sub-components of the tracker. The performance score (AUC value) for each tracker is shown in the legend. Bottom: Tracking results shown on a frame from the ``Diver'' sequence when using the baseline, baseline$+$NetC and the full algorithm. 
}
\label{FigSP}
\end{figure}

In this section, we describe how we evaluated the contributions of the key components of the proposed method (i.e., NetC, NetT branches and adaptation) to the overall performance. In the first experiment, we designed our baseline algorithm to only apply the shared network NetS which fed into one branch of NetT. Since NetC was not used to classify the tracked category, the branch of pedestrian category in NetT was manually chosen as the pedestrian category dominates the test dataset. Note that the baseline algorithm fine-tunes NetT based on the initial bounding box. In the second experiment, we combined the baseline model with NetC to activate the corresponding (category-based) branch in NetT. In this stage, we also adapt the triggered NetT in the first frame while no inter-supervised adaptation takes place between the networks during tracking. This experiment shows how much the semantic (category) information can improve the performance. Finally, we performed the experiment with enabled inter-supervision between NetC and NetT to observe further improvements of the performance, as shown in Fig.~\ref{FigSP}. 

It is interesting to note that the baseline algorithm which uses the pedestrian branch of the NetT network for all testing videos ($64\%$ of the sequences, in fact, belong to other, non-pedestrian categories) still shows a relatively strong performance. For example, despite using a non-optimal NetT branch (i.e., pedestrian) for most of the sequences, it still performs favourably compared to DST~\cite{Tracking2016Xiao} (0.498, ranked 6th) and DSST~\cite{danelljan2014accurate} (0.475, ranked 7th) in the overall evaluation. This relatively strong performance can be attributed to the NetT fine-tuning initialisation step which adapts the branch for a particular tracking video. Adding NetC to the baseline algorithm results in significant improvements, which demonstrates the effectiveness of the semantic information. This can also be observed in Fig.~\ref{FigSP} (bottom), which shows that for a deforming target, the baseline tracker gradually drifts to the background while both NetC enhanced baseline algorithm and the full algorithm can track the diver robustly. The adaptation process, by inter-supervision between NetC and NetT further advances the overall performance (shown in Fig.~\ref{FigSP} plots).

\subsection{Qualitative comparison among CNN-based trackers}\label{SecCNNcompare}

\begin{figure*}[htb]
\centering
\includegraphics[width = 0.9 \textwidth]{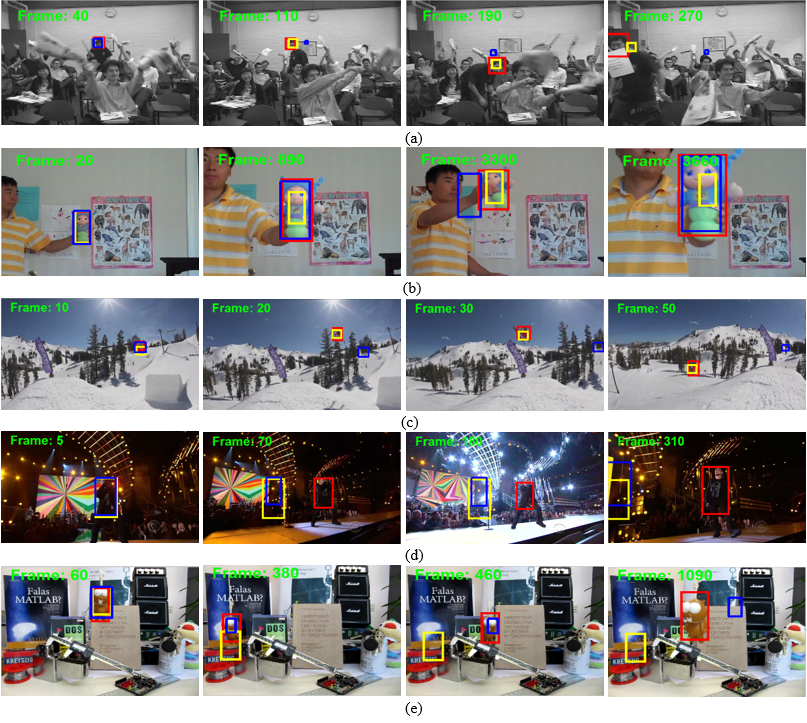}
\caption{Qualitative results of the CNN based trackers (~\textcolor{red}{red: ours};~\textcolor{yellow}{yellow: HCF}~\cite{Ma2015ICCV};~\textcolor{blue}{blue: DLT}~\cite{wang2013learning}.) on sequences: (a)~\textit{freeman4}; (b)~\textit{doll}; (c)~\textit{skiing}; (d)~\textit{singer2}; (e)~\textit{lemming}. }
\label{FigCNN}
\end{figure*}

We compare our tracker to other methods~\cite{Ma2015ICCV,wang2013learning} which also have the same major component, i.e., CNN, as our proposed semantic tracker. Ma et al.~\cite{Ma2015ICCV} utilised the pre-trained VGG model (from ImageNet) as a feature extractor, together with the kernelised correlation filter tracking framework. Since HCF tracker~\cite{Ma2015ICCV} only utilised the off-line trained model, a comparison between our work and HCF demonstrates the effectiveness of the online learning part for the proposed tracker. Note that the scale of HCF tracker~\cite{Ma2015ICCV} is not adapted, thus this comparison also shows the advantages of applying the bounding box adaptation for our tracker. Different from HCF~\cite{Ma2015ICCV}, Wang et al.~\cite{wang2013learning} utilised the CNN for online learning which also distinguished the foreground target from a background like our NetT network. A comparison to DLT ~\cite{wang2013learning} (its performance is shown in Fig.~\ref{FigCNN}) demonstrates a superior performance of our tracker due to the semantic information and inter-supervised network adaptation jointly from NetC and NetT.

In the sequences containing objects with significant scale variations, e.g. \textit{freeman4}, \textit{doll}, HCF~\cite{Ma2015ICCV} tracker fails in tracking the object accurately. This is because HCF tracker cannot adapt the scale of the template. In contrast, our approach which implements scale adaptation can successfully deal with this problem. Note that HCF still exhibits the advantage of using the sophisticated features learned from ImageNet in the sequence \textit{skiing}, compared to the online trained DLT~\cite{wang2013learning} tracker. This is because  DTL tracker online trains the network purely based on the tracking results without additional supervision. When the target appearance changes dramatically, e.g., significant illumination conditions in sequence \textit{singer2} and a partial occlusion in \textit{lemming}, DTL tracker will gradually learn the background information and incorporate it into the model which will finally result in a failure. In contrast, our tracker benefits from the semantic knowledge about the target category, which provides more reliable training data to update the network in a robust way.

\subsection{Overall performance comparison}\label{SecOPC}
We evaluated our proposed tracker by comparing it to 29 original trackers in OTB~\cite{wu2015object} and additional 9 recently published trackers, namely, CCT~\cite{zhu2015collaborative}, LCT~\cite{ma2015long}, KCF~\cite{henriques2015high}, MEEM~\cite{zhang2014meem}, DSST~\cite{danelljan2014accurate}, TGPR~\cite{gao2014transfer}, DST~\cite{Tracking2016Xiao}, and CNN-based trackers HCF~\cite{Ma2015ICCV} and DLT~\cite{wang2013learning}. The AUC score of the top 10 trackers in terms of the success plots are shown in Tab.~\ref{TabSuccessPlot}, which shows the results obtained on 1) the whole dataset and 2) sub-datasets annotated with specific attributes, i.e., deformation (39~sequences), scale variation (61~sequences), illumination variation (35~sequences), low resolution (9~sequences), out-of-view (14~sequences) and fast motion (37~sequences). As shown in Tab.~\ref{TabSuccessPlot}, the proposed \textit{semantic tracker} outperforms all other 38 state-of-the-art trackers, not only overall, but also on the sub-datasets annotated with specific attributes, namely IV, SV, DEF, FM, OV and LR. 

\begin{table*}[htb]
\centering
\caption{The AUC score of OPE~\cite{wu2015object} success plots for the top 10 compared trackers. The best tracker is in bold, while the second best is denoted with *. IV: illumination variation; OPR: out-of-plane rotation; SV: scale variation; OCC: occlusion; DEF: deformation; MB: motion blur; FM: fast motion: IPR: in plane rotation; OV: out of view; BC: background clutter; LR: low resolution.}
\label{TabSuccessPlot}
\begin{tabular}{|l||llllllllllll|}
\hline
       & Overall & IV & OPR & SV & OCC & DEF & MB & FM & IPR & OV & BC & LR \\\hline\hline
Ours   & \textbf{0.572} &\textbf{0.577}&0.544*&\textbf{0.562}&0.514&\textbf{0.573}&0.577*&\textbf{0.589}&0.532&\textbf{0.516}&0.516&\textbf{0.568}\\\hline
HCF    & 0.562*&0.540&0.531&0.486&\textbf{0.520}&0.525*&\textbf{0.585}&0.578*&\textbf{0.559}&0.474&\textbf{0.585}&0.388 \\\hline
LCT    & 0.562&0.556*&\textbf{0.547}&0.500*&0.515*&0.507&0.533&0.560&0.557*&0.452&0.550*&0.399\\\hline
CCT    &0.549&0.533&0.519&0.486&0.482&0.514&0.541&0.571&0.516&0.430&0.521&0.432*\\\hline
MEEM   &0.530&0.521&0.530&0.479&0.512&0.496&0.556&0.557&0.529&0.488*&0.519&0.382\\\hline
DST   &0.498 &0.456&0.461&0.402&0.461&0.498&0.512&0.513&0.491&0.437&0.487&0.318\\\hline
DSST   & 0.475 &0.486&0.465&0.414&0.446&0.433&0.467&0.452&0.485&0.374&0.477&0.314\\\hline
KCF    &0.475 &0.474&0.463&0.396&0.456&0.455&0.459&0.465&0.465&0.393&0.497&0.290\\\hline
Struck & 0.459&0.430&0.435&0.406&0.405&0.403&0.456&0.469&0.447&0.359&0.427&0.313\\\hline
TGPR   & 0.458&0.448&0.456&0.405&0.430&0.460&0.429&0.421&0.462&0.373&0.428&0.344\\\hline
\end{tabular}
\end{table*}

\subsection{Failure cases}

\begin{figure*}[!htb]
\centering
\includegraphics[width = 0.9 \textwidth]{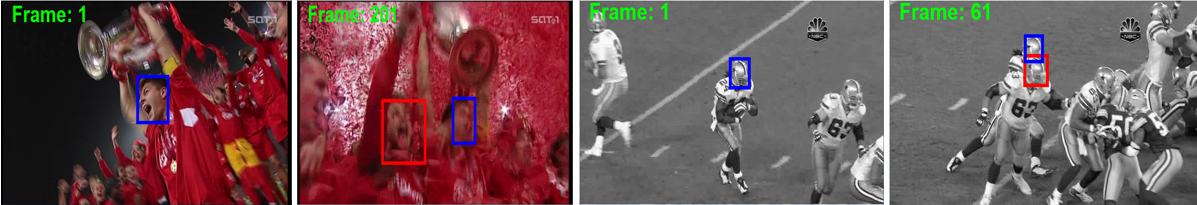}
\caption{Two examples of failure cases of the semantic tracker. The blue bounding boxes indicate the (annotated) ground truth, while the red bounding boxes were output by our semantic tracker.}
\label{FigFC}
\end{figure*}

It is also important to identify and analyse the failure cases of the designed algorithm. We show two such examples in Fig.~\ref{FigFC}. Even though our tracker has achieved superior performance both overall and on the sub-sequences with annotated attributes, it still has difficulties tracking  objects in scenes with camouflage. In such cases, semantic information only about the target itself is not sufficient to distinguish the object from the background which has an identical appearance as the target. To tackle these types of problems, the tracker should also exploit the semantic information contained in the scene~\cite{Tracking2016Xiao}.

\section{Conclusions}\label{SecCon}

In this paper, we proposed a new single target {\em semantic tracker} which intertwines the processes of target classification and target tracking. This is achieved by a novel network structure which comprises of different CNNs, i.e., a shared convolutional network (NetS), a classification network (NetC) and a tracking network (NetT). These networks are trained to encompass both generic features and category-specific features. During online tracking, consistent outputs of NetC and NetT jointly determine the sample regions with the right category and foreground labels for target estimation, while inconsistencies in the outputs of NetC and NetT trigger adaptation of the networks. The extensive experiments have shown that our tracker outperforms 38 state-of-the-art tracking algorithms tested on a large scale tracking benchmark OTB~\cite{wu2015object} with 100 sequences. Note that our current work only considers the semantic information of the objects, and that a lack of contextual semantic information may cause tracking difficulties/failures in highly cluttered scenes or when tracking objects without distinguishing features, such as translucent objects, as mentioned in~\cite{Tracking2016Xiao}. Therefore, in future, we will also exploit contextual semantic information and improve the performance of the tracker in cases of camouflage. In addition, our ongoing work will also focus on scaling up the proposed semantic tracker to a larger number of categories. This requires the tracker to construct multi-branches of NetT network in a more automatic, self-organised, way.

\section*{Acknowledgement}
We acknowledge MoD/Dstl and EPSRC for providing the grant to support the UK academics (Ales Leonardis) involvement in a Department of Defense funded MURI project. This work was also supported by EU H2020 RoMaNS 645582 and EPSRC EP/M026477/1.

\medskip

{\small
\bibliographystyle{abbrv}
\bibliography{egbib}
}

\begin{IEEEbiography}
[{\includegraphics[width=1in,height=1.75in,clip,keepaspectratio]{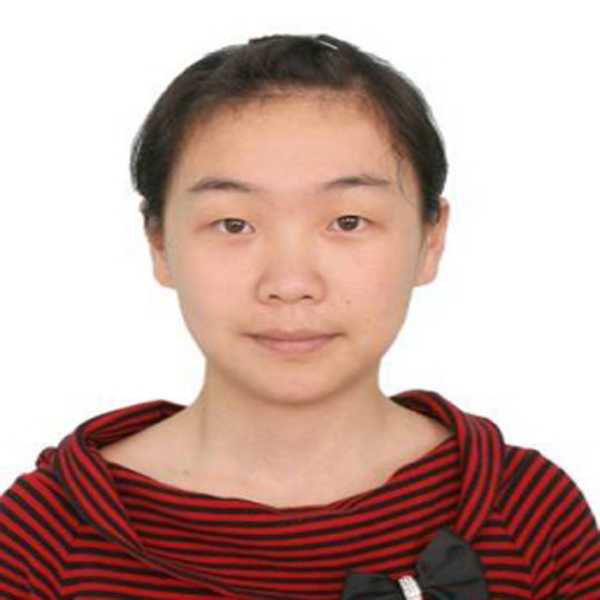}}]{Jingjing Xiao} received her Bachelor and
Master degree from College of Mechatronics Engineering and
Automation, National University of Defence Technology, China, in 2010
and 2012, respectively. She holds a PhD degree from University of Birmingham in 2016. She received the best poster award in the BMVA summer school in 2014. Currently, she is a research fellow in the University of Birmingham, U.K..Her research interests include single object tracking, multi-object tracking with computer vision.
\end{IEEEbiography}

\begin{IEEEbiography}
[{\includegraphics[width=1in,height=1.25in,clip,keepaspectratio]{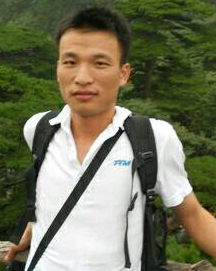}}]{Qiang Lan} received the M.S. and B.S. degrees in computer science from the National University of Defense Technology (NUDT), Changsha, China. He continues his PhD degree in computer science and technology in NUDT. His research topics are about high performance computing and computation optimization in  convolutional neural network.
\end{IEEEbiography}

\begin{IEEEbiography}
[{\includegraphics[width=1in,height=1.5in,clip,keepaspectratio]{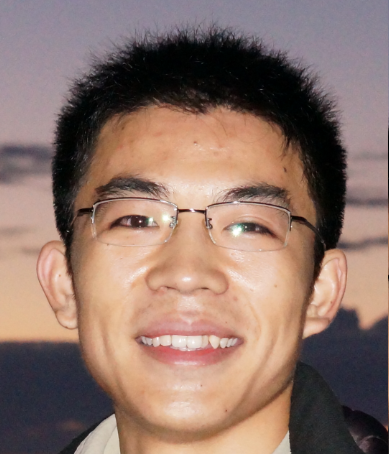}}]{Linbo Qiao} received the M.S. and B.S. degrees in computer science from the National University of Defense Technology (NUDT), Changsha, China, where he is currently pursuing the PhD degree in computer science and technology. His research interests include machine learning, online and distributed computation. 
\end{IEEEbiography}

\begin{IEEEbiography}
[{\includegraphics[width=1in,height=1.25in,clip,keepaspectratio]{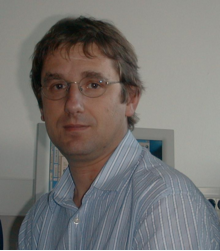}}]{Ale\v{s} Leonardis} is Professor at the School of Computer Science, University of Birmingham and co-Director of the Centre for Computational Neuroscience and Cognitive Robotics. He is also Professor at the FCIS, University of Ljubljana and adjunct professor at the FCS, TU-Graz. His research interests include robust and adaptive methods for computer vision, object and scene recognition and categorization, statistical visual learning, 3D object modelling, and biologically motivated vision.
\end{IEEEbiography}

\end{document}